\pdfoutput=1

\documentclass[11pt]{article}

\usepackage{EMNLP2023}

\usepackage{times}
\usepackage{latexsym}
\usepackage{float}

\usepackage[utf8]{inputenc}
\usepackage{tikz}
\usetikzlibrary{trees}

\usepackage{tikz}
\usetikzlibrary{positioning,shapes.geometric}

\usepackage[T1]{fontenc}

\usepackage[utf8]{inputenc}

\usepackage{microtype}

\usepackage{inconsolata}
\usepackage{graphicx}
\usepackage{booktabs}
\usepackage{amssymb} 
\usepackage{tabularx} 

\usepackage{listings}
\usepackage{xcolor}
\usepackage[normalem]{ulem}
\lstdefinelanguage{yaml}{
  keywords={true,false,null,y,n},
  keywordstyle=\color{blue}\bfseries,
  basicstyle=\ttfamily\footnotesize,
  morecomment=[l]{\#},
  commentstyle=\color{gray}\ttfamily,
  stringstyle=\color{orange}\ttfamily,
  morestring=[b]',
  morestring=[b]"
}
\lstdefinestyle{yamlstyle}{
  backgroundcolor=\color{lightgray},
  basicstyle=\ttfamily\footnotesize,
  frame=single,
  tabsize=2,
  breaklines=true,
  showstringspaces=false
}
\def\MEdel#1{\bgroup\markoverwith{\textcolor{red}{\rule[0.5ex]{2pt}{1pt}}}\ULon{#1}}

\title{Domain Adaptation of Llama3-70B-Instruct through Continual Pre-Training and Model Merging: A Comprehensive Evaluation}

\author{Shamane Siriwardhana, Mark McQuade, Thomas Gauthier, Lucas Atkins, Fernando Fernandes Neto\\
{\bf Luke Meyers, Anneketh Vij, Tyler Odenthal, Charles Goddard, Mary MacCarthy, Jacob Solawetz  } \\
Arcee, Florida, USA \\
\texttt{\{shamane, mark, thomas, lucas, fernando, luke, anneketh, tyler, charles, mary, jacob\}@arcee.ai}}

\begin{document}
\maketitle

\begin{abstract}
We conducted extensive experiments on domain adaptation of the Meta-Llama-3-70B-Instruct model on SEC data, exploring its performance on both general and domain-specific benchmarks. Our focus included continual pre-training (CPT) and model merging, aiming to enhance the model's domain-specific capabilities while mitigating catastrophic forgetting. Through this study, we evaluated the impact of integrating financial regulatory data into a robust language model and examined the effectiveness of our model merging techniques in preserving and improving the model's instructive abilities. The
model is accessible at hugging face \href{https://huggingface.co/arcee-ai/Llama-3-SEC-Base}{arcee-ai/Llama-3-SEC-Base}. This is an intermediate checkpoint of our final model, which has seen 20B tokens so far. The full model is still in the process of training.

\textbf{This is a preprint technical report with thorough evaluations to understand the entire process}.
\end{abstract}

\section{Introduction}
In the rapidly-evolving landscape of language models, training models in specific domains has become increasingly crucial. Domain-specific large language models (LLMs) are fine-tuned to understand and generate text within particular fields, significantly enhancing their performance and relevance \cite{lin2024flame, gekhman2024does}. This specialized training process allows these models to leverage vast amounts of domain-specific data to improve their accuracy and utility.

Continual Pre-Training (CPT) and domain adaptation for large language models (LLMs) are the core techniques employed today for enhancing the performance and domain specificity of LLMs. The Domain-specific Continual Pre-Training \cite{que2024dcpt} focuses on refining LLMs by optimizing the amalgamated ratio of general corpus data to domain-specific data. This approach has been demonstrated to effectively predict performance across different mixture ratios and dataset sizes with minimal training costs, thereby improving the adaptability and efficiency of LLMs in various domains.

The base model for our work, Meta-Llama-3-70B-Instruct \cite{llama3modelcard}, has significantly impacted the LLM community with its robust capabilities and adaptability. It serves as a solid foundation for our domain-specific enhancements, particularly in integrating SEC data to create a specialized chat agent.

SEC (Securities and Exchange Commission) data encompasses a wide range of financial and regulatory information submitted by publicly-traded companies. This data includes quarterly and annual reports, filings related to insider trading, proxy statements, and other critical documents. SEC data is important for investors, analysts, and regulators as it provides transparent and detailed insights into a company's financial health, operational performance, and corporate governance.
The SEC Chat Agent is utilized in several key areas. To mention a few: Investors rely on SEC data to make informed decisions by analyzing financial statements, earnings reports, and market disclosures. Financial institutions and analysts make use of the SEC filings to assess companies' risk profiles, thereby improving risk mitigation strategies. Regulators and compliance officers ensure adherence to financial and operational regulations by identifying potential violations through SEC data. Researchers and policymakers examine SEC filings to understand corporate governance practices, promoting transparency and accountability. Additionally, market analysts and strategists track industry trends, competitive positioning, and market opportunities using SEC data.

Our main contributions in this technical report 
\begin{itemize}
    \item \textbf{Large-Scale Continual Pre-Training (CPT) of Llama-70B-Instruct on SEC Data:} We conducted extensive CPT with 70 billion tokens of SEC data, significantly enhancing the model’s understanding and generation capabilities within the financial domain.
    
    \item \textbf{Comprehensive Evaluation of Domain-Specific Performance:} Our experiments demonstrated substantial improvements in domain-specific tasks such as financial classification and numerical reasoning, highlighting the effectiveness of CPT in adapting the model to specialized content.
    
    \item \textbf{Effective Mitigation of Catastrophic Forgetting through Model Merging:} By employing TIES merging techniques, we observed that up to certain levels, we could bring back the general capabilities of the original instruct model while enhancing domain-specific knowledge, thereby addressing catastrophic forgetting and maintaining a balance between specialized and general performance.
    
    \item \textbf{Future Directions for Enhancing Model Capabilities:} We highlight future directions to further restore original capabilities, such as post-merge supervised fine-tuning (SFT), direct preference optimization (DPO), and better alignment techniques.
\end{itemize}

\section{Background}
The continuous advancements in the domain of natural language processing (NLP), the Large language models (LLMs), brings up the need for domain adapted models. Models which are optimized for the specific tasks, are able to provide clear, concise and correct information to the user prompt is the much needed use case, Both from business's perspective and for the end users. This paper explores how domain adaptation \cite{gururangan2020dont} of LLM models can be enriched using the continual pre-training \cite{ibrahim2024simple} combined with Model Merging \cite{goddard2024arcees} techniques and it's applications.The domain adaption has an array of application areas, this paper explores the Llama3-70B-Instruct domain adaption with the U.S. Securities and Exchange Commission (SEC) data.

\subsection{Domain adaptation of LLMs}
Domain adaptation of models plays an important role in the fine-tuning of pre-trained LLM models. Data being the key element in the domain adaptation, with the increase in number of domain specific tokens, corpus of highly structured data, model capabilities emerges \cite{wei2022emergent} to better understand domain specific semantics, syntax and the sentiment of the data. The increased efficiency, lowered perplexity and proven accuracy allows for the model to be deployed in mission critical applications. We explore domain adaptation \cite{liu2024chipnemo} techniques to improve LLM model performance on niche domain specific tasks.

\subsection{Continual Pre-Training}
In language, CPT was studied under the name of domain adaptation pre-training where the new dataset comes from a new domain \cite{gupta2023continual}. For instance, PMC-LLaMA \cite{wu2023pmcllama}, an open-source medical-specific large language model, incorporates data-centric knowledge injection with pure CPT and medical-specific instruction tuning. It stands out as the first of its kind, showcasing superior performance on diverse medical benchmarks with significantly fewer parameters compared to both ChatGPT and LLaMA-2 \cite{touvron2023llama}. As another example, ChipNeMo \cite{liu2024chipnemo} investigates the utility of large language models (LLMs) in industrial chip design, employing a domain-adaptive CPT approach in their adaptation process. They assess their model across three specific chip design applications: an engineering assistant chatbot, EDA script generation, and bug summarization and analysis. Their findings demonstrate that their domain adaptation pipeline enhances LLM performance substantially compared to general-purpose models, achieving up to a 5x reduction in model size while maintaining or improving performance across various design tasks.[3] Inspired by prior work, CPT at Arcee involves extending the training of a base model, such as Llama-2-base \cite{touvron2023llama} or Mistral-7B-base \cite{jiang2023mistral}, using domain-specific datasets. This process allows us to fine-tune models to the nuances of specialized fields.

\subsection{Catastrophic Forgetting}
Domain adaptation is paramount at Arcee AI, yet traditional methodologies demand considerable time and resources. A significant challenge arises with catastrophic forgetting \cite{Kirkpatrick_2017}, wherein post-pretraining often results in a deterioration of the model's original general abilities hindering its fine-tuned performance across various tasks. This underscores the need for a method capable of incorporating domain-specific knowledge while mitigating forgetting and other deterioration. Our breakthrough lies in integrating two key methodologies: Continual Pre-Training (CPT) and Model Merging, designed to enhance efficiency and efficacy in adapting language models to specific domains to overcome the challenges like catastrophic forgetting.

\subsection{Model Merging}
Model Merging involves synthesizing the capabilities of multiple pre-trained models into a single, more versatile checkpoint. This technique enables us to combine domain-specific models with general-purpose chat models, leveraging the strengths of both.\cite{yadav2023tiesmerging, stoica2024zipit}. Leveraging MergeKit \cite{goddard2024arcees}, We explored various merging techniques, such as Linear \cite{wortsman2022model} SLERP \cite{Digitous2024}, TIES \cite{yadav2023tiesmerging} and DARE \cite{yu2024language} to integrate our CPT checkpoints with general-purpose chat models. Model Merging maintains a balance between general and domain-specific knowledge while mitigating the risk of catastrophic forgetting, as the weights in the foundational general model can remain frozen. This stage was crucial for enhancing the model's adaptability and performance in specific domains. 

\section{Data Acquisition and Pre-Processing}
Our process for reading, parsing, extracting, cleaning, and storing pure text data from SEC filings involves several key steps. We use libraries like boto3 for AWS S3 interactions, trafilatura for text extraction, and various components from the Hugging Face's DataTrove \cite{Penedo_DataTrove_large_scale} library. 

First, we read data from AWS S3, specifically targeting .txt files. HTML documents are processed using trafilatura to extract text and convert it to Markdown format. This ensures easier processing and analysis. We then filter documents based on acceptable formats, such as Markdown and plaintext, while avoiding non-textual content like PDFs, Excel files, and ZIP archives for model training.

Datatrove processing pipelines are used to streamline data handling, enabling efficient processing of large volumes of SEC filings by executing stages such as reading, cleaning, filtering, and writing concurrently. Finally, the cleaned files are saved back to the S3 bucket, organized under a specified prefix for easy access and further analysis. This comprehensive preprocessing pipeline ensures high-quality input for our model's training.

\section{Continual Pre-Training (CPT) with Megatron-Core}
Training a model at the scale of 70 billion parameters presents significant challenges in terms of efficiency and scalability. While many training frameworks exist, they often fall short when handling such large models and token counts. To address this, we opted for Megatron \cite{shoeybi2020megatronlm}, leveraging its advanced parallelization capabilities, including model parallelism, tensor parallelism, and sequence parallelism.

Our training was conducted on a cutting-edge AWS SageMaker HyperPod cluster, consisting of 4 nodes, each equipped with 32 H100 GPUs. This robust setup was essential for the efficient and scalable training of our massive dataset. Usually, SLURM clusters are particularly challenging to set up but are crucial for our needs. However, SageMaker HyperPod was easy to configure and had already been tested with Megatron, making it an ideal choice for our training environment. SageMaker HyperPod ensures streamlined distributed training for large clusters, optimizing the utilization of compute, memory, and network resources. It provides a resilient training environment by repairing faulty instances. Its FSx drive ensures faster saving, and the InfiniBand network facilitates faster communication. Additionally, it can integrate with dashboards like Grafana for better monitoring. We integrated everything into the Arcee Training Platform to create a seamless training experience.
In the CPT layer, we mixed 70B token SEC data with a general sample of Together AI’s RedPijama data of 1B tokens by following prior work \cite{ibrahim2024simple}. This approach has allowed us to maintain a high level of generalization while benefiting from domain-specific data. The total estimated time to train using Megatron on this distributed training setup was 31 days, highlighting the significant computational effort involved.

\begin{figure}[h]
  \centering
  \includegraphics[width=\linewidth]{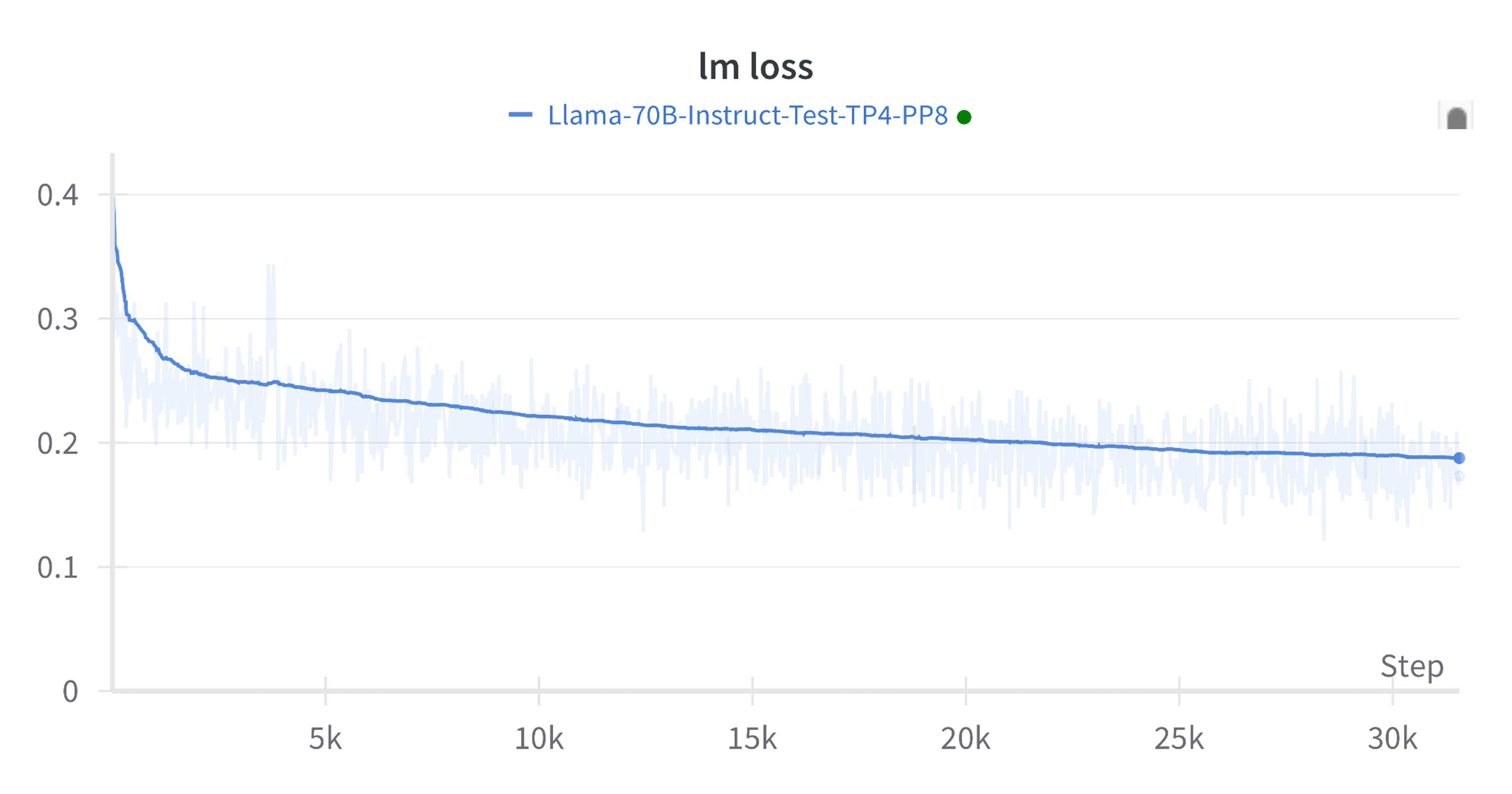}
  \caption{LM Loss}
  \label{fig:1}
\end{figure}

\begin{figure}[h]
  \centering
  \includegraphics[width=\linewidth]{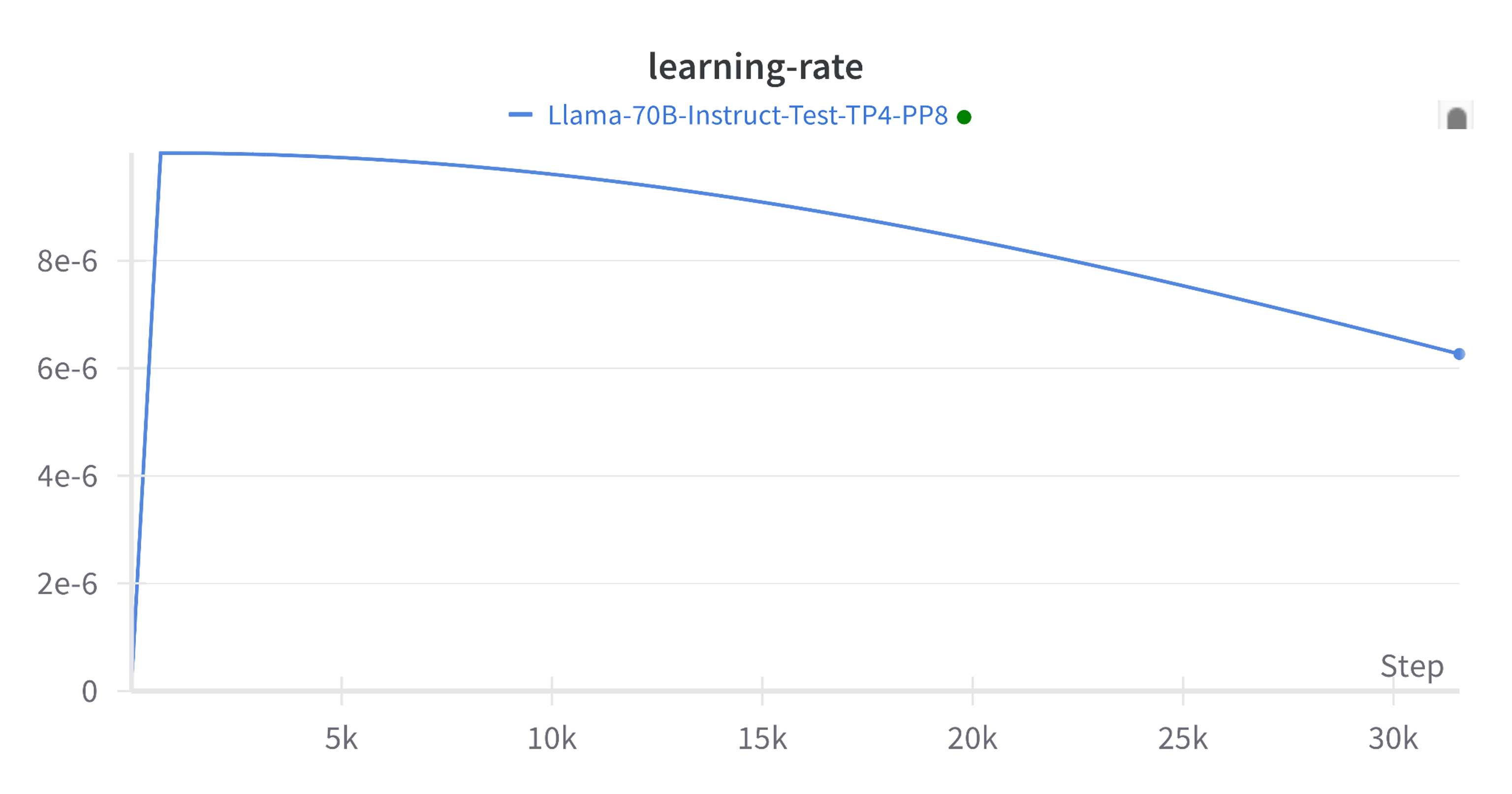}
  \caption{Learning Rate}
  \label{fig:2}
\end{figure}

During our training, we initially processed 70B tokens. However, in this release, we are sharing the CPT model trained with 20B tokens, since the 70B model is still being trained. We plan to release additional checkpoints in the future. Figure~\ref{fig:1} and Figure~\ref{fig:2} illustrate the current LM loss and the learning rate curves.

\section{Post CPT Model Merging with MergeKit}
To mitigate the issue of catastrophic forgetting that can occur during continual pre-training, we employed TIES merging \cite{yadav2023tiesmerging} from the Arcee Mergekit \cite{goddard2024arcees} toolkit. Drawing on prior work~\cite{labrak2024biomistral}, we merged the CPT models back into the instruct version. The primary goal of the merge was to retain the instructive capabilities of the base model while integrating the specialized knowledge acquired during CPT. 

TIES merging is a natural fit here and allows us to maintain a balance between the instruct model's foundational understanding and the CPT model’s enhanced domain-specific expertise, resulting in a more robust and versatile language model. The merge config is shown in the Listing~\ref{lst:yaml-merge-config}.

\begin{lstlisting}[language=yaml, caption={Merge Configuration}, label={lst:yaml-merge-config}]
merge_method: ties
base_model: meta-llama/Meta-Llama-3-70B
models:
  - model: /home/ubuntu/data/cpt
    parameters:
      weight:
        - filter: mlp
          value: [0.25, 0.5, 0.5, 0.25]
        - filter: self_attn
          value: [0.25, 0.5, 0.5, 0]
        - value: [0.25, 0.5, 0.5, 0.25]
      density: 0.75
  - model: meta-llama/Meta-Llama-3-70B-Instruct
    parameters:
      weight:
        - filter: mlp
          value: [0.75, 0.5, 0.5, 0.75]
        - filter: self_attn
          value: [0.75, 0.5, 0.5, 1]
        - value: [0.75, 0.5, 0.5, 0.75]
      density: 1.0
parameters:
  normalize: true
  int8_mask: true
dtype: bfloat16
\end{lstlisting}

\section{Evaluation}
To ensure the robustness of our model, we conducted thorough evaluations on both domain-specific and general benchmarks. Domain-specific evaluations are crucial to assess our model's performance within its targeted domain. However, general evaluations are equally important to ensure no catastrophic forgetting of the model's original capabilities.
In every evaluation, we compared the following models with each other:
\begin{enumerate}
\item \textbf{Llama-70B-Instruct (meta-llama/Meta-Llama-3-70B-Instruct)}: The original instruct model released by Meta.
\item \textbf{Llama-70B-CPT}: The Llama-70B-Instruct model after continual pre-training, with the checkpoint saved after seeing 20B tokens.
\item \textbf{Llama-70B-CPT-Merge}: The Llama-70B-CPT model merged with the original Llama-70B-Instruct model using the TIES method.
\end{enumerate}

\subsection{Domain-Specific Evaluations Metrics}
Domain-specific perplexity is crucial for evaluating a model's performance within its targeted domain, ensuring effective adaptation to domain data. Tracking perplexity changes helps assess the impact of continual pre-training and domain-specific improvements.

\begin{figure}[h]
  \centering
  \includegraphics[width=\linewidth]{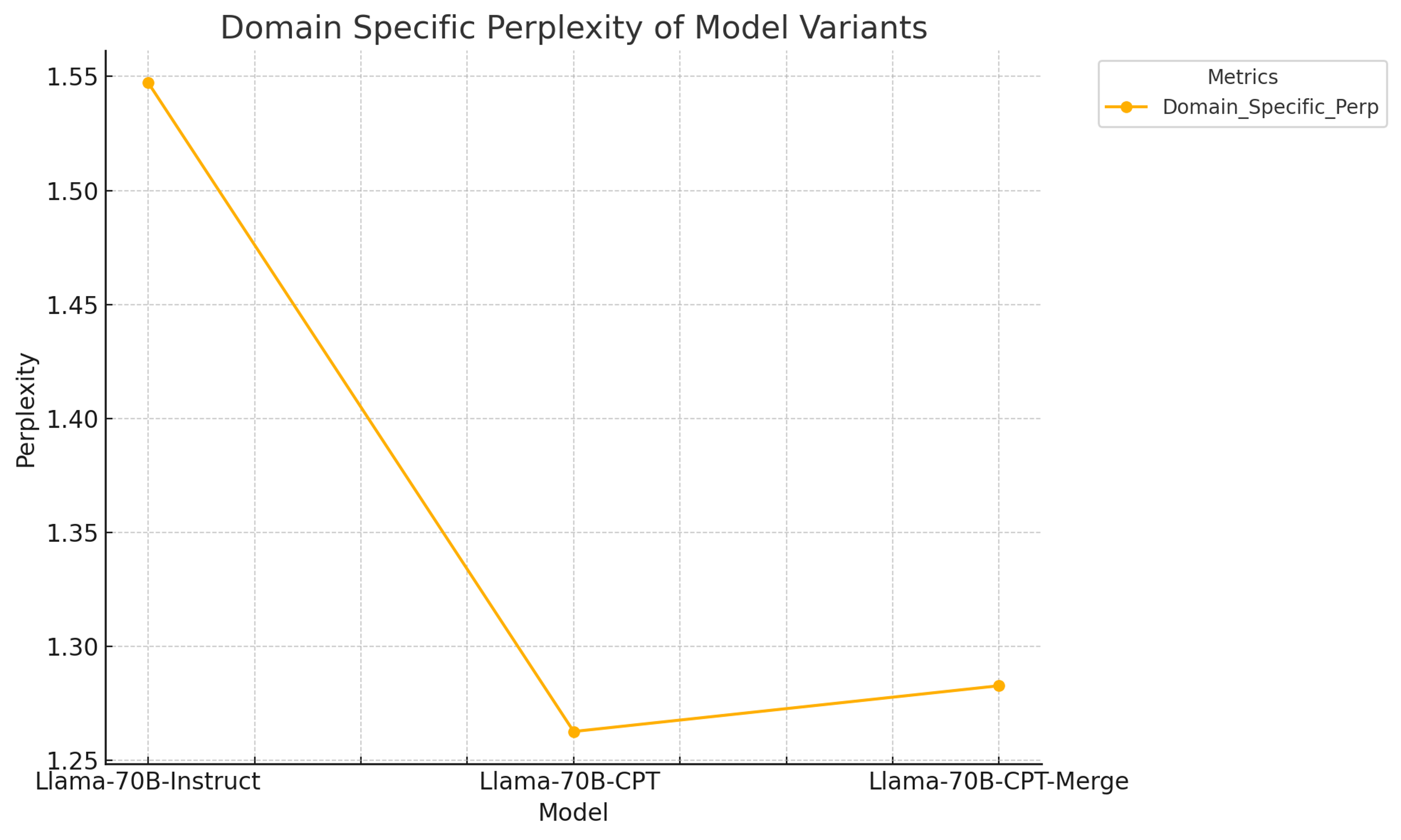}
  \caption{Domain Specific Perplexity of Model Variants (lower the better)}
  \label{fig:3}
\end{figure}

Figure~\ref{fig:3} illustrates domain-specific perplexity measurements for different model variants, highlighting the impact of continual pre-training and model merging on SEC data performance.

\begin{figure}[h]
  \centering
  \includegraphics[width=\linewidth]{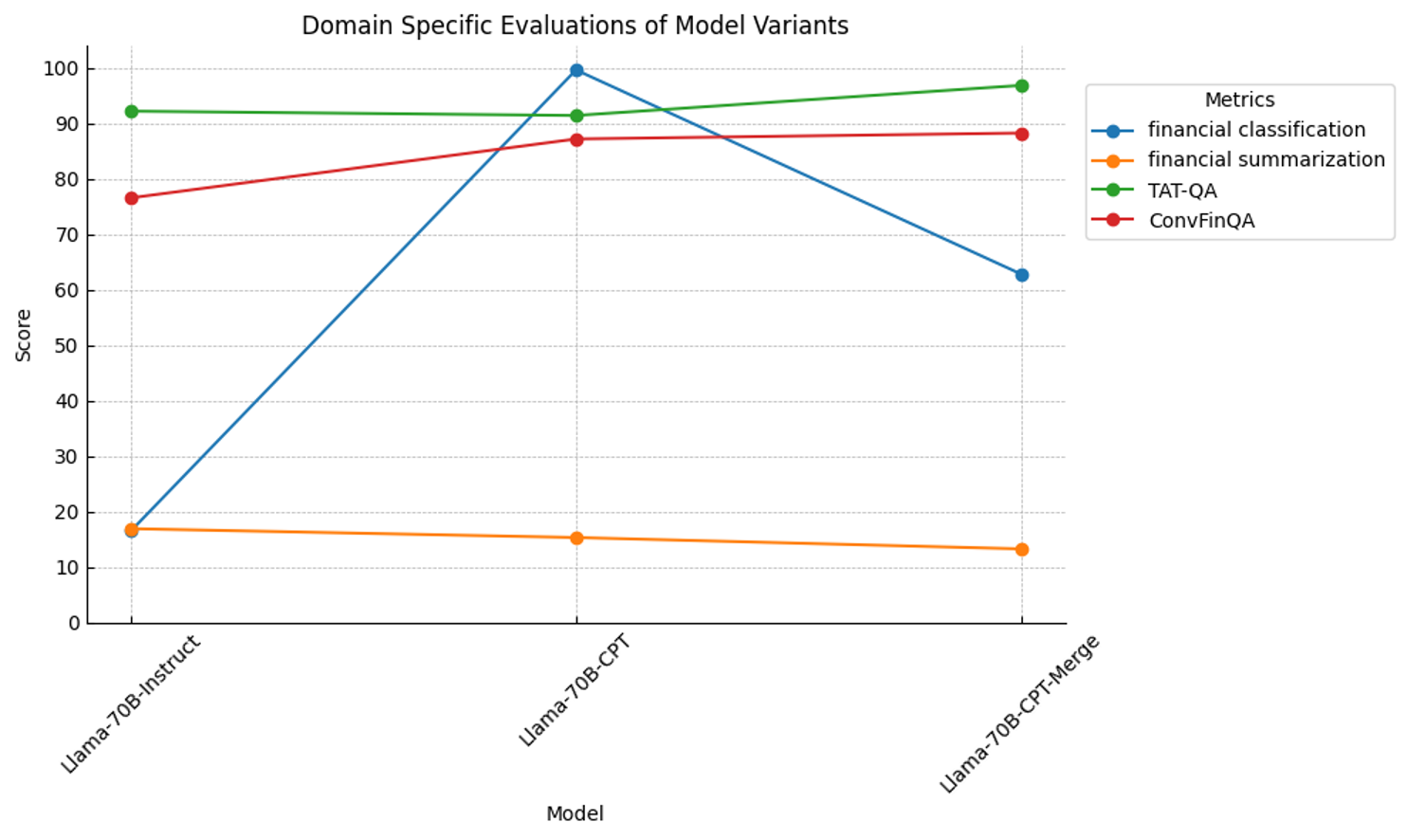}
  \caption{Domain Specific Evaluations of Model Variants}
  \label{fig:4}
\end{figure}

Insights Figure~\ref{fig:3} : 
\begin{itemize}
\item CPT reduces perplexity related to SEC data, indicating the model's improved understanding and adaptation to this specific domain.

\item Merging the CPT model with the Llama3-Instruct version increases perplexity slightly, likely due to reintroducing some of the lost chat capabilities.

\item Despite a slight increase in perplexity post-merging, the final model maintains a lower perplexity compared to the original, demonstrating effective domain adaptation while retaining chat capabilities. This indicates that merging models does not compromise the infused domain knowledge gained during continual pre-training. As illustrated in Figures~\ref{fig:4}, \ref{fig:5}, and \ref{fig:6}, model merging can enhance instruction-following capabilities. This flexibility is invaluable for our objectives, as it allows us to combine domain-specific expertise with improved instruction-following abilities without sacrificing performance.
\end{itemize}

As illustrated in Figure~\ref{fig:4}, for domain-specific evaluations, we test the model's performance on extractive numerical reasoning tasks, namely a subset of TAT-QA \cite{zhu-etal-2021-tat} and ConvFinQA \cite{chen2022convfinqa} which are not precisely related to SEC data but still relevant for evaluating domain-specific performance.

Insights Figure~\ref{fig:4} : 

\begin{itemize}
\item For ConvFinQA, there is a clear improvement in performance after CPT and further improvement after merging with the instruct model.
\item For TAT-QA, significant improvement is observed only after merging, likely due to its specialization in hybrid tabular and textual content, which is less represented in SEC data.
\item For the financial classification, where the model categorizes texts as premises or claims, we see very significant accuracy improvements after CPT, nearing a perfect score and indicating that the model learns new tasks effectively from the unsupervised training on SEC data. Merging loses some accuracy but still sits very comfortably above the Instruct baseline.
\item For the financial text summarization task, the consistent ROUGE-1 scores across all checkpoints suggest that training on SEC data does not improve performance, possibly due to the baseline model's already strong capabilities and the inherent limitations of ROUGE, which relies on potentially imperfect reference summaries.
\item These results highlight the importance of merging models to recover general capabilities, demonstrating how merging can enhance performance in specialized tasks by reintroducing broader knowledge and capabilities.
\end{itemize}

\subsection{General Evaluations Metrics}

The following Figure~\ref{fig:5} illustrates the performance comparison, based on general evaluation metrics using the Eval Harness \cite{eval-harness} in full precision.

We focused on an updated version of the Nous research benchmark consisting of the following metrics: 

\begin{itemize}
    \item BIG-bench \cite{srivastava2023imitation}
    \item AGIEval \cite{zhong2023agieval}
    \item GPT4all: A Combination of HellaSwag, OpenBookQA, Winogrande, ARC Easy, ARC Challenge, BoolQ, PIQA)
    \item TruthfulQA \cite{lin2022truthfulqa}
\end{itemize}

\begin{figure}[h]
  \centering
  \includegraphics[width=\linewidth]{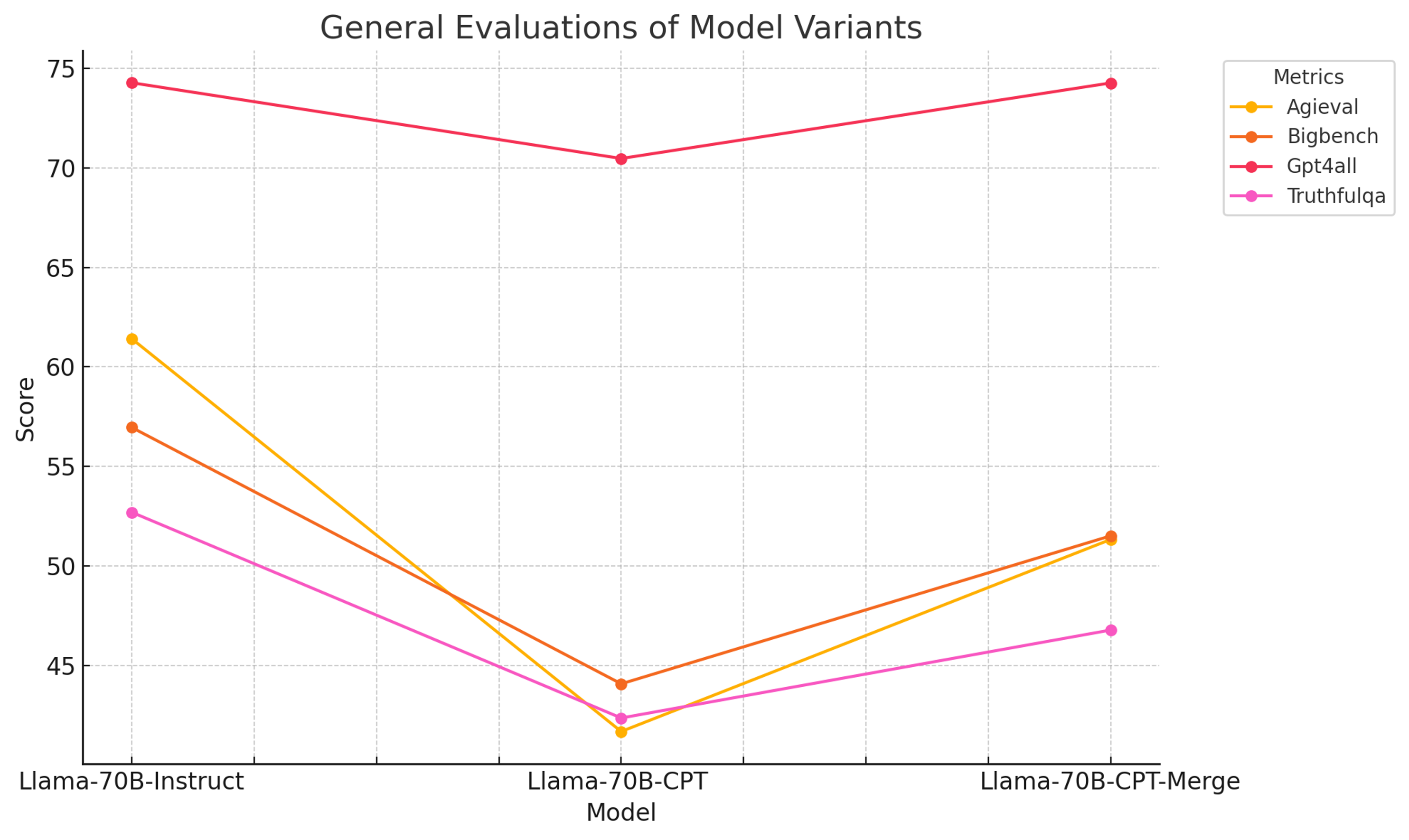}
  \caption{General Evaluations of Model Variants}
  \label{fig:5}
\end{figure}

Insights Figure~\ref{fig:5}:
\begin{itemize}
\item CPT of Llama-70B-Instruct, resulting in Llama-70B-CPT, demonstrates a drop in general evaluation scores across all metrics (AGIEval, BigBench, GPT4all, TruthfulQA), indicating potential catastrophic forgetting.
\item The drop in performance is most prominent in GPT4all and AGIEval metrics, highlighting the challenge of maintaining general capabilities while adapting to new domains.
\item Merging Llama-70B-CPT with the original Llama-70B-Instruct model using the TIES method (resulting in Llama-70B-CPT-Merge) significantly recovers the lost general capabilities.
\item This recovery is evident across all metrics, suggesting that model merging can effectively mitigate the catastrophic forgetting observed during continual pre-training.
\item These findings underscore the importance of model merging in maintaining a balance between domain adaptation and general capabilities, making it a valuable technique for continual pre-training processes.
\end{itemize}

In the context of CPT, it’s crucial to measure general perplexity to evaluate the model's performance. It’s important to know how well the model can work with the previous knowledge. Perplexity measures on the following general datasets are used:
\begin{enumerate}
    \item bigcode/starcoderdata \cite{starcoderdata}
    \item open-web-math/open-web-math \cite{paster2023openwebmath}
    \item allenai/peS2o \cite{peS2o}
    \item mattymchen/refinedweb-3m \cite{refinedweb3m}
    \item Wikitext \cite{merity2016pointer}
\end{enumerate}

\begin{figure}[h]
  \centering
  \includegraphics[width=\linewidth]{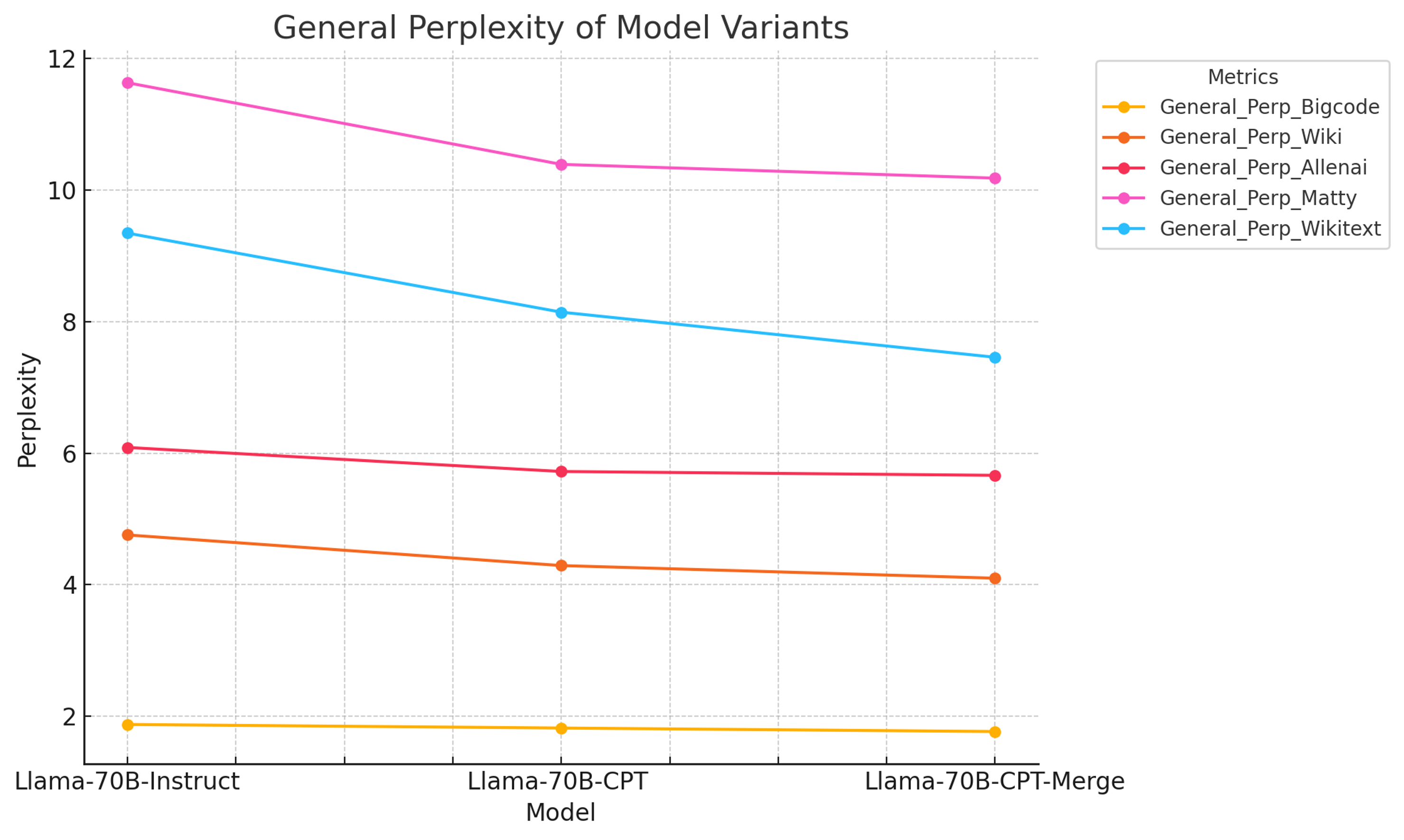}
  \caption{General Perplexity of Model Variants}
  \label{fig:6}
\end{figure}

Insights Figure~\ref{fig:6}: 

\begin{itemize}
 \item CPT with a substantial amount of SEC domain-specific tokens (20B) reduces perplexity across all general datasets, indicating improved predictive capabilities.
   \begin{itemize}
    \item This could be due to the nature of the SEC data.
  \end{itemize}
 \item The model maintains familiarity with general domains even after extensive domain-specific training, as indicated by the stable perplexity metrics for general text shown in the graph. This demonstrates that CPT does not degrade the model's general knowledge, although it may reduce certain capabilities. As seen in Figure~\ref{fig:3}, we propose that future work on better SFT adaptations can help restore the model's instruction following capabilities, leveraging the retained knowledge.
\end{itemize}

\section{Conclusion and Future Work}
This marks the beginning of a series of model releases by Arcee AI focused on developing the best domain-specific chat models, as well as showcasing the power of training paired with model merging. In this release, we’re sharing the CPT model trained with 20B tokens, since the final model is still being trained. We plan to release additional checkpoints in the future.

Aligning CPT Models with SFT, DPO, and other alignment methods 
Addressing catastrophic forgetting by aligning CPT models with general Supervised Fine-Tuning (SFT), Direct Preference Optimization (DPO), and Reinforcement Learning from Human Feedback (RLHF). While model merging can recover some lost knowledge, integrating these alignment techniques will further minimize catastrophic forgetting and help realize the next version of our models.

Improving the CPT Data Processing Layer
Enhancing data filtering methods to better manage catastrophic forgetting and optimize data mixing with general data, crucial for handling large-scale models like the 70B.

Exploring Model Merging Further
Investigating advanced techniques and methodologies for model merging to maximize the retention of general capabilities while enhancing domain-specific performance.

Finally, we invite the community to explore these techniques and contribute to the ongoing efforts in alignment and merging, fostering collaborative advancements in the development of domain-specific chat models.

\bibliography{custom}
\bibliographystyle{acl_natbib}




\end{document}